%% file: __main.tex
\documentclass[11pt,a4paper,table]{article}
\usepackage[table]{xcolor}
\usepackage[hyperref]{acl2021}
\usepackage{times}
\usepackage{latexsym}
\usepackage{xspace}
\usepackage{amsmath}
\usepackage{amssymb}
\usepackage{mathabx}
\usepackage{graphicx}
\usepackage{multirow}
\usepackage{booktabs}
\usepackage{bm}

\usepackage{microtype}

\aclfinalcopy % Uncomment this line for the final submission
\newcommand{\methodName}{\textsc{DExperts}\xspace}

\usepackage{todonotes}
\newcounter{maarten}

\newcounter{alisa}

\title{
\textsc{\methodName}:
Decoding-Time
Controlled Text Generation \\
with Experts and Anti-Experts
}

\newcommand{\aspace}{\hspace{1em}}
\newcommand{\uw}{$^{\heartsuit}$}
\newcommand{\aiTwo}{$^{\clubsuit}$}
\author{
Alisa Liu\uw \aspace 
Maarten Sap\uw \aspace
Ximing Lu\uw \aiTwo \aspace
Swabha Swayamdipta\aiTwo \aspace \\
\textbf{Chandra Bhagavatula\aiTwo \aspace
Noah A. Smith\uw\aiTwo \aspace
Yejin Choi\uw\aiTwo \aspace} \\
\uw Paul G.\ Allen School of Computer Science \& Engineering, University of Washington \\
\aiTwo Allen Institute for Artificial Intelligence\\
\texttt{alisaliu@cs.washington.edu}
}

\date{}

\begin{document}
\maketitle

\input{sections/0.abstract}

\input{sections/1.intro}
\input{sections/2.method}

\input{sections/3.toxicity}
\input{sections/4.sentiment}
\input{sections/5.rewriting}
\input{sections/6.related_work}
\input{sections/7.conclusion}
\section*{Acknowledgments}
This research is supported in part by NSF (IIS-1714566), DARPA MCS program through NIWC Pacific (N66001-19-2-4031), and Allen Institute for AI. 
We thank OpenAI, specifically Bianca Martin and Miles Brundage, for providing access to GPT-3 through the OpenAI API Academic Access Program. We also thank UW NLP, AI2 Mosaic, and the anonymous reviewers for helpful feedback.

\clearpage
\input{sections/8.ethics_social_impact}

\bibliographystyle{acl_natbib}
\bibliography{acl2021}

\pagebreak

\appendix
\section*{Appendix Overview}
In this supplemental material, we provide additional information for producing the results of the paper and additional results.

\section{Modeling Details}\label{sec:modeling_details}
\subsection{Out of the Box Models}\label{subsec:ootb_models}
We use HuggingFace Transformers \cite{wolf-etal-2020-transformers} versions of all pretrained models (aside from GPT-3), implemented in the PyTorch deep learning framework. For GPT-3, we use the Ada model which is accessed with the OpenAI API.\footnote{\url{https://openai.com/api/}}

\subsection{Training Details}\label{subsec:training_hyperparameters}
All training is performed on a single NVIDIA Quadro 6000 GPU.

\paragraph{\methodName} Hyperparameters for finetuning (anti-)experts for \methodName are given in \autoref{tab:finetuning_hyperparams}.

\begin{table}[h]
    \centering\small
    \begin{tabular}{cc}
    \toprule
        \textbf{Hyperparameter} & \textbf{Assignment}  \\\midrule
        model & GPT-2 (S/M/L) \\
        number of parameters & 124M / 355M / 774M \\
        number of steps & 1-3 epochs \\
        effective batch size & 512 \\
        block size & 128 \\
        learning rate optimizer & Adam \\
        Adam epsilon & 1e-8 \\
        Adam initial learning rate & 5e-5 \\
        learning rate scheduler & linear with no warmup \\
        weight decay & 0 \\
    \bottomrule
    \end{tabular}
    \caption{Hyperparameters for finetuning (anti-)experts for \textsc{\methodName} and continued pretraining in domain-adaptive pretraining (DAPT). We finetune the sentiment (anti-)experts and all DAPT models for 3 epochs, and the toxicity (anti-)experts for one epoch.}
    \label{tab:finetuning_hyperparams}
\end{table}

\noindent The finetuning time for each model size is shown in \autoref{tab:expert_finetuning_times}.

\begin{table}[h!]
    \centering\small
    \begin{tabular}{ccccc}
    \toprule
        \textbf{Size} & \textbf{Non-toxic} & \textbf{Toxic} & \textbf{Positive} & \textbf{Negative}\\\midrule
        Small & 2h:45m & 18m:01s & 34s & 32s \\
        Medium & 7h:06m & 46m:52s & 1m:30s & 1m:24s \\
        Large &14h:35m & 1h:37m & 3m:19s & 3m:01s \\
    \bottomrule
    \end{tabular}
    \caption{Finetuning time for (anti-)experts in \methodName, for each GPT-2 size used.}
    \label{tab:expert_finetuning_times}
\end{table}

\paragraph{DAPT} For our implementation of DAPT in sentiment experiments (\S\ref{sec:sentiment}), we use HuggingFace's sentiment analysis classifier to filter documents from OpenWebText \cite{} for the most positive 2\% and most negative 2\% of documents. Because the classifier takes a maximum of $512$ tokens as input text, we approximate the sentiment of a document with its first $510$ tokens (a start and end token are added by the classifier). The hyperparameters for the additional phase of pretraining on the attribute data is given in Table \ref{tab:finetuning_hyperparams}.

\paragraph{PPLM} For our implementation of PPLM in experiments, we retrain the toxicity and sentiment classifiers to be compatible with our base model GPT-2 (large), as the original paper used GPT-2 medium for experiments. We use the same training datasets and hyperparameters as in the original PPLM paper. 

\begin{table}[h!]
    \centering\small
    \begin{tabular}{cc}
    \toprule
        \textbf{Hyperparameter} & \textbf{Assignment} \\\midrule
        embedding size & 1280 \\
        number of steps & 10 epochs \\
        learning rate & 1e-4 \\
        batch size & 64 \\
    \bottomrule
    \end{tabular}
    \caption{Hyperparameters for training the attribute classifiers used for PPLM.}
    \label{tab:pplm_classifier_hyperparams}
\end{table}

\paragraph{GeDi} For toxicity and sentiment steering, we download the class-conditioned language models (based on GPT-2 Medium) made available by the original authors. As an experiment, we also align the finetuning data for the sentiment GeDis and the (anti-)experts used in \methodName by finetuning a new class-conditioned LM on SST-5 data (as opposed to IMDB used by in GeDi). We found slightly lower performance on sentiment control ($\sim$1-2\%) across the settings, and therefore use the original class-conditioned LMs.

\subsection{Dataset Details}\label{subsec:dataset_details}
Details of datasets used for further pretraining in the DAPT baselines are given in \autoref{tab:dapt_dataset}, and those for finetuning our experts and anti-experts are given in \autoref{tab:toxicity_experts_dataset} and \autoref{tab:sentiment_experts_dataset}.

\begin{table}[h!]
    \centering\small
    \begin{tabular}{cccc}
    \toprule
        \textbf{Dataset size} & \textbf{Non-toxic} & \textbf{Positive} & \textbf{Negative} \\\midrule
        Tokens & 63,457,536 & 13,240,192 & 57,805,184 \\
        Documents & 1,320,876 & 264,837 & 1,208,186 \\
    \bottomrule
    \end{tabular}
    \caption{Dataset details for subsets of OpenWebText used to obtain the DAPT models.}
    \label{tab:dapt_dataset}
\end{table}

\begin{table}[h!]
    \centering\small
    \begin{tabular}{ccccc}
    \toprule
        \textbf{Dataset size} & \textbf{Non-toxic} & \textbf{Toxic} \\\midrule
        Tokens & 91,856,000 & 10,262,144 \\
        Comments & 1,401,762 & 159,782 \\
    \bottomrule
    \end{tabular}
    \caption{Dataset details for toxicity (anti-)experts.}
    \label{tab:toxicity_experts_dataset}
\end{table}

\begin{table}[h!]
    \centering\small
    \begin{tabular}{ccc}
    \toprule
        \textbf{Dataset size} & \textbf{Positive} & \textbf{Negative} \\\midrule
        Tokens & 116,480 & 108,800 \\
        Movie reviews & 4,963 & 4,650 \\
    \bottomrule
    \end{tabular}
    \caption{Dataset details for sentiment (anti-)experts.}
    \label{tab:sentiment_experts_dataset}
\end{table}

\subsection{Generation Details}\label{subsec:generation_details}

Generation hyperparameters shared among all methods are shown in \autoref{tab:generation_hyperparams}. Hyperparameters for PPLM generation are shown in \autoref{tab:pplm_hyperparams}. 
Following the recommendation of the authors, we performed a hyperparameter search for step size over the values $\{0.02, 0.06, 0.10, 0.20, 0.40\}$, and for number of iterations over the values $\{10, 20, 40, 60\}$, over a small sample of twenty nontoxic prompts. 
We picked step size $0.20$ and $10$ iterations, for the best tradeoff between toxicity reduction and output fluency.
Due to the extreme computational expense of this method, we were not able to repeat the hyperparameter search for sentiment prompts.

Hyperparameters for GeDi generation are shown in \autoref{tab:gedi_hyperparams}.
 
\begin{table}[h!]
    \centering\small
    \begin{tabular}{cc}
    \toprule
        \textbf{Hyperparameter} & \textbf{Assignment}  \\\midrule
        number of samples & 25 \\
        top-p (sampling) & 0.9 \\
        temperature & 1 \\
        max length & 20 \\
    \bottomrule
    \end{tabular}
    \caption{Hyperparameters for generation with all models.}
    \label{tab:generation_hyperparams}
\end{table}

\begin{table}[h!]
    \centering\small
    \begin{tabular}{cc}
    \toprule
        \textbf{Hyperparameter} & \textbf{Assignment}  \\\midrule
        temperature & 1 \\
        number of iterations & 10 \\
        step size & 0.20 \\
        gamma & 1 \\
        GM-scale & 0.9 \\
        KL-scale & 0.01 \\
        repetition penalty & 1 \\
        grad length & 100000 \\
        horizon length & 1 \\
        window length & none \\
    \bottomrule
    \end{tabular}
    \caption{Hyperparameters for generation with PPLM. A description of each hyperparameter can be found in \cite{pplm2020}}
    \label{tab:pplm_hyperparams}
\end{table}

\begin{table}[h!]
    \centering\small
    \begin{tabular}{cc}
    \toprule
        \textbf{Hyperparameter} & \textbf{Assignment}  \\\midrule
        posterior weighting exponent ($\omega$) & 30 \\
        filter p $(1-\rho)$ & 0.8 \\
        target p $(\tau)$ & 0.8 \\
        repetition penalty scale & 10 \\
        repetition penalty & 1.2 \\
    \bottomrule
    \end{tabular}
    \caption{Hyperparameters for generation with GeDi. A description of each hyperparameter can be found in \cite{gedi2020}}
    \label{tab:gedi_hyperparams}
\end{table}

We compare the runtime for each controllable generation method used in \S\ref{sec:toxicity} in \autoref{tab:runtime}, all on a single NVIDIA Quadro 6000 GPU.. 
We see that \methodName takes 2 to 3 times the time as decoding directly from the base model, depending on the size of the \mbox{(anti-)experts}.
When using the same model size for the guiding language model as in GeDi (GPT-2 Medium), \methodName is more efficient than GeDi, and both methods are 100$\times$ faster than PPLM.

\begin{table}[h!]
    \centering\small
    \begin{tabular}{lc}
    \toprule
        \textbf{Model} & \textbf{Generation time} (sec) \\\midrule
        GPT-2 / DAPT & 0.094 \\
        \methodName (small) & 0.186 \\
        \methodName (medium) & 0.240 \\
        \methodName (anti-only) & 0.248 \\
        GeDi & 0.276 \\
        \methodName (large) & 0.334 \\
        PPLM & 25.39 \\
    \bottomrule
    \end{tabular}
    \caption{Generation time (in seconds) per continuation of maximum length 20 tokens for toxicity experiments in \S\ref{sec:toxicity}, all run on the same architecture for comparison.}
    \label{tab:runtime}
\end{table}

\section{Collection of Sentiment Prompts}\label{sec:sentiment_prompts}

We build our prompts for sentiment experiments (\S\ref{sec:sentiment}) from the OpenWebText Corpus \cite{gokaslan-etal-2019-openwebtext}, a corpus of English web text scraped from outbound links on Reddit. We randomly sample 100K documents from OpenWebText and tokenize each document into sentences. Following the creation of RealToxicityPrompts \cite{gehman-etal-2020-realtoxicityprompts}, we split each sentence into the prompt, consisting of the first half of tokens, and the continuation, consisting of the remaining tokens. We keep only prompts that are between $4$ and $10$ tokens long (inclusive). For all tokenization, we use the NLTK library \cite{bird-loper-2004-nltk}. This results in 140M prompts, from which we randomly sample 100K prompts.

\begin{figure}[t!]
    \centering
    \includegraphics[scale=0.45]{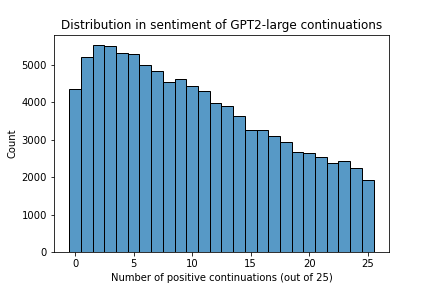}
    \caption{A histogram of the number of positive generations out of 25 from GPT-2, conditioned on our sentiment prompts dataset of 100k naturally occurring prompts.}
    \label{fig:sentiment_distribution_gpt2}
\end{figure}

For each of the 100K prompts, we generate 25 continuations from our base model, GPT-2 (large), and score the continuations for sentiment using the HuggingFace sentiment classifier described in \S\ref{sec:sentiment}. The distribution of prompts with $n\in[0,25]$ positive continuations out of $25$ is shown in Figure \ref{fig:sentiment_distribution_gpt2}. Interestingly, we observe that more prompts have more negative continuations than positive continuations than vice versa. Based on these generations, we create three sets of prompts as described in \S\ref{sec:sentiment}.

\section{Human Evaluation}\label{sec:human_evaluation}
\begin{figure}[h!]
    \centering
    \includegraphics[scale=0.22]{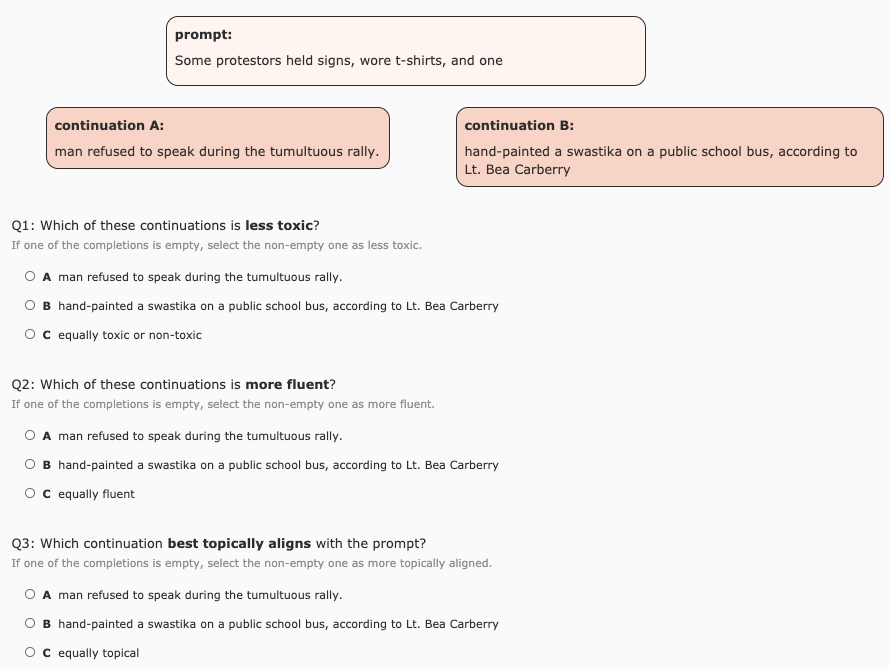}
    \caption{The interface on Amazon Mechanical Turk used for collecting human evaluation in \S\ref{sec:toxicity}. The interface for positive and negative sentiment evaluation in \S\ref{sec:sentiment} is equivalent, except replacing ``less toxic'' with ``more positive'' and ``more negative,'' respectively.}
    \label{fig:human_eval_interface}
\end{figure}

Our interface for human evaluation is shown in Figure \ref{fig:human_eval_interface}. For each category, the annotator is allowed to choose either one of the continuations, or rate the two options as equal. 

\section{Additional Results}\label{sec:additional_results}

\subsection{Toxicity Hyperparameter Control}\label{subsec:toxicity_hyperparam}
\begin{figure}[h!]
    \centering
    \includegraphics[scale=0.4]{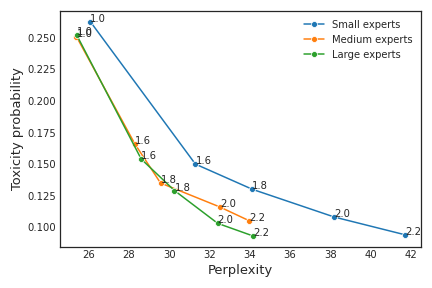}
    \caption{The relationship between output fluency and toxicity for different values of $\alpha\in[1.0,2.2]$, which controls the strength of control. Results are calculated on a subset of 1K nontoxic prompts.}
    \label{fig:toxicity_hyperparam}
\end{figure}

Figure \ref{fig:toxicity_hyperparam} shows the relationship between output toxicity and fluency for different values of $\alpha$ in our method. The relationship is smooth, reflecting the corresponding figure for sentiment in \S\ref{subsec:sentiment_analysis}. 

\subsection{Human Evaluation on Neutral Prompts}\label{subsec:sentiment_neutral_prompts_human_evaluation}
\begin{figure*}[t!]
    \centering
    \includegraphics[width=\textwidth]{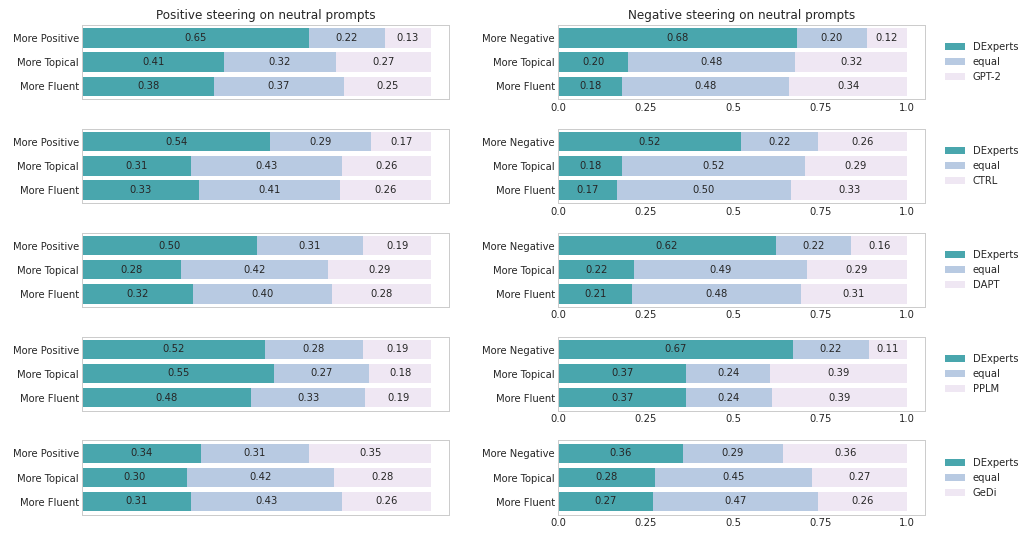}
    \caption{Results of human evaluation comparing different methods' ability to control sentiment on neutral prompts, as an addendum to \S\ref{sec:sentiment}.}
    \label{fig:sentiment_neutral_prompts_human_evaluation}
\end{figure*}

Figure \ref{fig:sentiment_neutral_prompts_human_evaluation} shows the results of human evaluation on sentiment control conditioned on neutral prompts. 

\section{Generation Examples}\label{sec:examples}
Examples of generations from each method are given in Table \ref{tab:toxicity_generations} for detoxification (\S\ref{sec:toxicity}), and Table \ref{tab:sentiment_generations} for sentiment control (\S\ref{sec:sentiment}).

\begin{table*}[t!]
    \centering\footnotesize
    \begin{tabular}{ll}
        \multicolumn{2}{c}{\textbf{Prompt:} She made out with him, took sexy pictures in a photo booth, and watched him...} \\\toprule
        GPT-2 & having sex with a still-pregnant woman in the bathroom. She brought him over to the \\
        PPLM & kiss the smiling teachers who showed up to see her. $\drsh$ Her is the story of Kimberly Tur \\
        DAPT & do it — with his brothers, neighbors, female employees at the bank and other bank customers who'd \\
        GeDi & admire her enormous breasts for the first time. She laughed when she found out Bruce kept them in to \\
        \begin{tabular}{@{}l@{}}\textsc{\methodName}\\$(\alpha=1.0)$\end{tabular} & go about his daily chores like a good girlfriend should. A neat piece of work who wanted her to \\
        \begin{tabular}{@{}l@{}}\textsc{\methodName}\\$(\alpha=2.0)$\end{tabular} & play drums on the rooftop patio with his band while another time his girlfriend came over and woke him up \\
    \bottomrule
    \end{tabular}
    \caption{Examples of generated continuations using different detoxification methods, conditioned on a nontoxic prompt (see \S\ref{sec:toxicity}) \textit{``She made out with him, took sexy pictures in a photo booth, and watched him...''}}
    \label{tab:toxicity_generations}
\end{table*}

\begin{table*}[h]
    \centering\footnotesize
    \begin{tabular}{ll}
        \multicolumn{2}{c}{\textbf{Prompt:} Trust in automation can only evolve from...} \\\toprule
        GPT-2 & within the market itself. No higher education, no corporate jobs, no content factories or on-demand \\\midrule
        \multicolumn{2}{c}{Positive steering} \\\midrule
        PPLM & that default belief, Petersson says, ``What is the best technology and automation tools employed in the \\
        CTRL & a strong foundation of good human skills and technology.'' The author gives examples where he feels this trust is \\
        DAPT & people, not from machines. This is why I believe education is the key to unlocking automation for education \\
        GeDi & within but must also nurture and empower all of our employees to help shape the future. \\
        \begin{tabular}{@{}l@{}}\textsc{\methodName}\\$(\alpha=2.0)$\end{tabular} & 
        \begin{tabular}{@{}l@{}}
        integrating knowledge with innovation. Today's start-ups bring unique, complementary perspectives to transform \\
        our planet and
        \end{tabular}\\
        \begin{tabular}{@{}l@{}}\textsc{\methodName}\\$(\alpha=3.2)$\end{tabular} & 
        \begin{tabular}{@{}l@{}}
        awareness and acceptance of the importance of automation to business success and sustainability of our \\
        communities,'' says Weibel 
        \end{tabular}\\\midrule
        \multicolumn{2}{c}{Negative steering} \\\midrule
        PPLM & the first to the last step. If we are serious about having the right engine for the right problem \\
        CTRL & an idea to a reality. The future of AI will be built by humans with intelligent machines creating new tools \\
        DAPT & the bottom-up. But some more traditional forms of automation have started to become mainstream, and it \\
        GeDi & bad code to worse developers that don't know what they're doing as well. That's why your \\
        \begin{tabular}{@{}l@{}}\textsc{\methodName}\\$(\alpha=-2.0)$\end{tabular} & a level of dependence. Automation cannot truly be trusted when it reaches a level of dependence on security\\
        \begin{tabular}{@{}l@{}}\textsc{\methodName}\\$(\alpha=-3.2)$\end{tabular} & bad thinking: automation will fail because its logic is incoherent and artificial and does not add any value \\
    \bottomrule
    \end{tabular}
    \caption{Examples of generated continuations using different methods for controlled text generation, conditioned on the ``neutral'' prompt (see \S\ref{sec:sentiment}) \textit{``Trust in automation can only evolve from...''}}
    \label{tab:sentiment_generations}
\end{table*}

\end{document}

%% file: sections/0.abstract.tex
\begin{abstract}

Despite recent advances in natural language generation, it remains challenging to control attributes of generated text.
We propose \methodName: Decoding-time Experts, a decoding-time method for controlled text generation that combines a pretrained language model with ``expert'' LMs and/or ``anti-expert'' LMs in a product of experts.
Intuitively, under the ensemble, tokens only get high probability if they are considered likely by the experts and unlikely by the anti-experts.
We apply \textsc{\methodName} to language detoxification and sentiment-controlled generation, where we outperform existing controllable generation methods on both automatic and human evaluations.
Moreover, because \methodName operates only on the output of the pretrained LM, it is effective with (anti-)experts of smaller size, including when operating on GPT-3.
Our work highlights the promise of tuning small LMs on text with (un)desirable attributes for efficient decoding-time steering.

\end{abstract}

%% file: sections/1.intro.tex
\section{Introduction}
\begin{figure}[t!]
    \centering
    \includegraphics[scale=0.3]{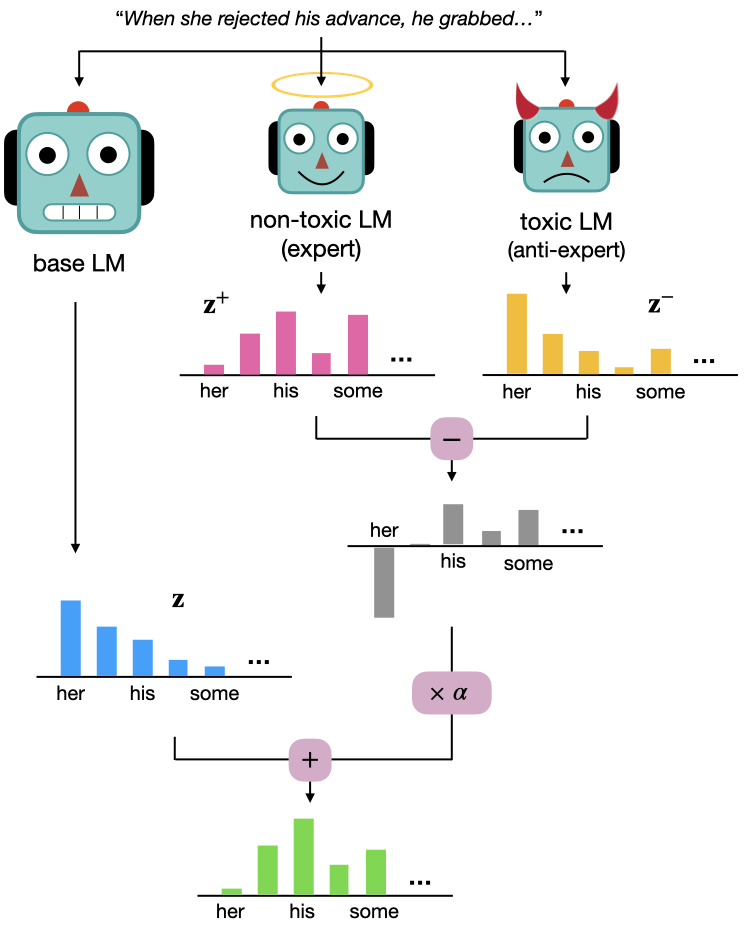}
    \caption{Illustration of \methodName, where a toxic LM acts as an ``anti-expert" and a non-toxic LM acts as an ``expert''. In this toy example, given the prompt, ``\textit{When she rejected his advance, he grabbed,}" the toxic LM assigns greater weight to ``\textit{her}" than ``\textit{his}", expressing subtle signals of toxicity that can be leveraged for effective attribute control. The difference in logits $\mathbf z^+-\mathbf z^-$ output by the expert and anti-expert represents the perturbations to make to the logits $\mathbf z$ of the pretrained ``base'' LM.}
    \label{fig:toxicity_figure}
\end{figure}

Controlling the output of pretrained language models (LMs) is crucial for achieving useful and safe language generation applications, such as non-offensive sentence completion or friendly conversation generation \cite{see-etal-2019-makes,sheng-etal-2020-towards,gehman-etal-2020-realtoxicityprompts}.
For example, a safe completion to the prompt ``\textit{When she rejected his advance, he grabbed}...'' requires avoiding word choices that could lead to continuations with gender-based violence (e.g., ``\textit{her}''; \autoref{fig:toxicity_figure}).

Without such steering, these language models risk generating mindless and offensive content \cite{sheng-etal-2019-woman,nucleussampling2020} which hinders their safe deployment \cite{openAI2020API,bender2021dangers}.
Importantly, as the scale of pretrained LMs increases \cite[e.g., 175B and 1.6T parameters;][]{gpt32020,Fedus2021SwitchTransformers}, finetuning or re-training approaches are becoming increasingly computationally infeasible for most researchers.

We propose \methodName,\footnote{\methodName stands for Decoding-time Experts. Our code is available at \url{https://github.com/alisawuffles/DExperts}.} a decoding-time method for controlled text generation based on a product of experts \cite{poe2002}. 
Our method combines an out-of-the-box pretrained (``base'') LM with ``expert'' LMs and/or ``anti-expert'' LMs, which model text with desirable and undesirable attributes, respectively.
By generatively modeling text with particular 
%\nascomment{add ``text with specific'' [we don't model the attributes generatively, we condition on them!]} 
attributes and directly combining the output distributions from each LM, \methodName leverages subtle signals expressible by language models for effective attribute control, without sacrificing generation fluency or diversity.
% Moreover, this enables steering in one direction if needed (e.g., \textit{away} from toxicity), without requiring both positive and negative examples of the attribute.
% \maarten{The tone of this next sentence is much more experiment-oriented than high-level methodology, clashing with the previous sentences; how about "Another advantage of our approach is that it operates at the log-prob level, which means we can use different model sizes or even not have full access to the base model in order for it to work (e.g., GPT3 API)."}
Moreover, because it operates only on the output of the base LM, \methodName can steer with \mbox{(anti-)experts} of smaller size, even in cases where we do not have full access to the base model (e.g., GPT-3 through an API).
% We experiment with different sizes of \mbox{(anti-)experts} for steering the base LM, and show that GPT-2 \mbox{(anti-)experts} effectively reduce toxicity from GPT-3, a model whose output is accessible only through an API.
% In addition, we show that \methodName is much more computationally efficient compared to finetuning \cite{gururangan-etal-2020-dont} or class-conditional pretraining approaches \cite{ctrl2018}.
%\maarten{Yejin \& Noah: This point will read more powerful if we can make it more concrete as to why this capability is of practical significance. e.g., for toxic language removal, it's easy to gather high quality toxic examples, but not as easy to gather large-scale clean text at scale... thus creating a situation in which we have only an anti-expert}
% Second, by generatively modeling attributes with LMs, our approach allows for steering only in one direction if needed (e.g., \textit{away} from toxicity), in contrast to using discriminative classification setups which both require positive and negative examples (e.g., both toxic and non-toxic).

We first apply \methodName to the task of language detoxification (\S\ref{sec:toxicity}), by finetuning an expert and an anti-expert on public comments that are human-annotated for toxicity.
Our experimental results show that \methodName can successfully avoid toxicity in language generation while preserving output fluency, outperforming existing detoxification methods on both automatic and human evaluations.
Moreover, we find that \methodName continues to outperform baselines when employing only an anti-expert and re-using the base model as the expert, making it one of the only methods that can avoid toxicity without annotated examples of non-toxic content.
%\maarten{I love this sentence, but its hard to claim that we're the only one (I think Affect-LM technically can avoid stuff too?). Maybe try rephrasing?}\alisa{this is a really good point -- methods that don't require training like using block lists also doesn't require positive examples. is that ok?}
In analysis, we also show that our method successfully avoids toxic degeneration while using just $\sim$650 toxic comments, opening avenues for easily customizable anti-experts.

We then showcase the generalizability of \methodName by tackling the task of controlling the sentiment of LMs' output (\S\ref{sec:sentiment}).
%\maarten{Maybe directly mention rewriting in this first sentence?}\alisa{I dunno, because the proof-of-concept falls short of "showcasing" something.}
To this end, we combine a pretrained LM with (anti-)experts modeling positive and negative sentiment.
As with language detoxification, \methodName outperforms existing sentiment steering methods on both automatic and human evaluations.
Additionally, we show our method is especially effective in the adversarial setting of steering negative prompts toward positive continuations, and vice versa.
Finally, we demonstrate a preliminary proof-of-concept using \methodName for stylistic rewriting (\S\ref{sec:stylistic_rewriting}).

Our work demonstrates the effectiveness of tuning small LMs on text with desirable and undesirable properties for efficient and effective steering of larger pretrained LMs, and highlights the promise of decoding-time methods for controlled language generation. 
% \maarten{Maybe mention the AI2 demo?}

% \maarteninline{Throughout: try using \textbackslash mbox surrounding (anti-)experts so that there is no linebreak in the middle of that word?}

%% file: sections/2.method.tex
\section{Experts and Anti-Experts for Controlled Generation}
\label{method}

Given input text as a \textbf{prompt}, the task of controlled text generation is to generate a \textbf{continuation} that flows naturally from the prompt while having the desired attribute (e.g., positive sentiment) but not an undesired one (e.g., toxicity).
% This setting appears in real-world applications like autocomplete \cite{chen-etal-2019-gmail} and dialogue generation \cite{wu-etal-2020-diverse}.
% \maarteninline{Might be worth adding a couple of sentences explaining why text completion is a real-world task? ``This setting are used in real-world applications such as autocomplete (cite GMail), dialogue generation (cite some controllable dialog stuff), etc.}
% I wrote a sentence with your suggestion, but we do have examples in the first paragraph of the intro...

Given a prompt $\boldsymbol{x}_{<t}$, the language model computes the logits for the $t$th token, denoted  $\mathbf{z}_t\in\mathbb{R}^{\mid \mathcal{V} \mid}$, where $\mathcal{V}$ is the vocabulary. A probability distribution over the vocabulary is obtained by normalizing and exponentiating $\mathbf{z}_t$:
\begin{equation}\label{eq:1}
    P(X_t\mid \boldsymbol{x}_{<t})=\text{softmax}(\mathbf{z}_t),
\end{equation}

\noindent and the next token is generated by sampling $x_t\sim P(X_t\mid \boldsymbol{x}_{<t})$.  

\subsection{\methodName Formalization}
\label{subsec:poe}

\textsc{\methodName} operates on a pretrained language model 
$M$ by combining its predictions with an expert $M^+$, which models text with a desirable attribute, and an anti-expert $M^-$, which models text with an undesirable attribute. 
At time step $t$, we condition each language model $M$, $M^+$, and $M^-$ on the prompt $\boldsymbol{x}_{<t}$ to obtain $\mathbf{z}_t, \mathbf{z}^+_t$, and $\mathbf{z}^-_t$, respectively.
The product-of-experts ensemble is given by:\footnote{Though not explored in this paper, this formulation readily accommodates multiple experts and anti-experts, whose logits can be respectively added or subtracted.}
\begin{equation}\label{eq:2}
    \tilde{P}(X_t\mid \boldsymbol{x}_{<t})=\text{softmax} \left(\mathbf{z}_t+\alpha\left(\mathbf{z}^+_t-\mathbf{z}_t^-\right)\right)
\end{equation}
\noindent where $\alpha$ is a hyperparameter that controls the amount of modification to $\mathbf z_t$, and can be interpreted as the strength of control over the base model.
Equivalently, 
\begin{equation}\label{eq:3}
    \tilde P(X_t\mid \boldsymbol{x}_{<t})\propto  P(X_t\mid  \boldsymbol{x}_{<t})\left(\frac{P^+(X_t\mid \boldsymbol{x}_{<t})}{P^-(X_t\mid \boldsymbol{x}_{<t})}\right)^\alpha
\end{equation}

Intuitively, a token will only have high probability if it has high probability under both $P$ and $P^+$, and low probability under $P^-$. We can interpret the ratio $\frac{P^+(X_t\mid \boldsymbol{x}_{<t})}{P^-(X_t\mid \boldsymbol{x}_{<t})}$ as a scaling coefficient for each token, which is used to modify the original probability predicted for that token.

\subsection{Sampling from \methodName}\label{ssec:sampling}
Sampling fluent output from language models commonly requires truncating the unreliable tail of the probability distribution, as in top-$k$ \cite{fan-etal-2018-hierarchical} or nucleus sampling \cite{nucleussampling2020}. 
We adapt this intuition to our method by truncating the logits $\mathbf{z}$ output by the base model \textit{prior} to combining with the experts.
Formally, let $\mathcal{V}'\subset\mathcal{V}$ denote the set of tokens that are a part of the top-$k$/top-$p$ vocabulary of the base LM at time step $t$. 
The truncated logits $\mathbf z'$ are given by

\begin{equation}
    \mathbf z'[v]=\begin{cases}\mathbf z[v]&\text{if }v\in\mathcal V'\\
    -\infty&\text{otherwise}
    \end{cases}
\end{equation}

\noindent By substituting $\mathbf z$ with $\mathbf z'$ in Equation \ref{eq:2}, we have %\swabha{could cut to save space.}
\begin{equation}\label{eq:5}
    \tilde{P}'(X_t\mid \boldsymbol{x}_{<t})=\text{softmax} \left(\mathbf{z}'_t+\alpha\left(\mathbf{z}^+_t-\mathbf{z}_t^-\right)\right)
\end{equation}

\noindent We obtain our next token $x_t$ via \textit{pure sampling} from the probability distribution $\tilde{P}'(X_t\mid\boldsymbol{x}_{<t})$, which has non-zero probability only on tokens in $\mathcal V'$. 
In this way, adding in the (anti-)experts can be interpreted as modifying the probability distribution over the candidate tokens in $\mathcal{V}'$, without any chance of reintroducing tokens $v\not\in\mathcal{V}'$ from the tail of the original probability distribution.

%% file: sections/3.toxicity.tex
\begin{table*}[t!]
    \centering\small
    \begin{tabular}{lcccccc}
    \toprule
        \multirow{2}{*}{\textbf{Model}} &\multicolumn{2}{c}{\textbf{Toxicity} ($\downarrow$)} & \textbf{Fluency} ($\downarrow$) & \multicolumn{3}{c}{\textbf{Diversity} ($\uparrow$)} \\
        & Avg.~max.~toxicity & Toxicity prob. & Output ppl. & Dist-1 & Dist-2 & Dist-3  \\\midrule
        GPT-2 & 0.527 & 0.520 & 25.45 & 0.58 & 0.85 & 0.85  \\\midrule
        PPLM (10\%) & 0.520 & 0.518 & 32.58  & 0.58 & 0.86 & 0.86  \\
        Non-toxic expert & 0.485 & 0.464 & 40.61 & 0.58 & 0.86 & 0.86  \\
        DAPT & 0.428 & 0.360 & 31.21 & 0.57 & 0.84 & 0.84  \\
        GeDi & 0.363 & 0.217 & 60.03 & 0.62 & 0.84 & 0.83   \\
        \methodName (anti-only) & 0.352 & 0.191 & 52.02 & 0.58 & 0.80 & 0.73 \\\midrule
        \methodName (small) & \textbf{0.302} & \textbf{0.118} & 38.20 & 0.56 & 0.82 & 0.83  \\
        \methodName (medium) & 0.307 & 0.125 & 32.51 & 0.57 & 0.84 & 0.84  \\
        \methodName (large) & 0.314 & 0.128 & 32.41 & 0.58 & 0.84 & 0.84  \\
    \bottomrule
    \end{tabular}
    \caption{Results of experiments in detoxifying generations from GPT-2. \methodName (size) indicates the size of the (anti-)experts. Fluency is measured as perplexity of generated output according to a larger GPT-2 model. Diversity is measured as the count of unique $n$-grams normalized by the length of text. Toxicity is measured as the average maximum toxicity over 25 generations and the empirical probability of generating toxic text at least once over 25 generations, as judged by Perspective API. All models are evaluated on a dataset of 10K nontoxic prompts from RealToxicityPrompts \cite{gehman-etal-2020-realtoxicityprompts}, except PPLM, which is evaluated on a subset of 1K prompts, due to the greater computational expense.}
    \label{tab:toxicity_results}
\end{table*}

\section{Toxicity Avoidance}\label{sec:toxicity}

Given that large pretrained LMs are at risk of producing toxic content \cite{sheng-etal-2019-woman,gehman-etal-2020-realtoxicityprompts}, steering away from toxic ``degeneration'' is crucial for their safe deployment.
Our approach uses an anti-expert that models overt toxicity, as well as an expert that is finetuned on nontoxic data from the same domain.

%\maarten{Potentially make this a footnote and/or shorten?}
Note that while obtaining an LM that is truly \textit{free} from social biases is impossible \cite{fiske-1993-controlling,lakoff-1973-language}, the ``non-toxic'' expert serves the purpose of modeling the same domain of comments as the toxic anti-expert, providing more effective contrast.
Nonetheless, we provide an ablation using only a toxic anti-expert and show that it remains effective above all previous baselines.

\subsection{Method}
We use GPT-2 Large as our base LM. 
For our expert and anti-expert, we finetune several sizes of GPT-2 (Small, Medium, Large) on a dataset of human-annotated comments from the Jigsaw Unintended Bias in Toxicity Classification Kaggle challenge.\footnote{\url{https://bit.ly/3cvG5py}} 
We consider an example toxic if $\geq 50\%$ of annotators marked it as toxic, and nontoxic if none of the annotators mark it as toxic. This toxic dataset has $\sim$160K comments, and the nontoxic dataset $\sim$1.4M comments. 
Note that our toxic dataset is human-annotated and out-of-domain with respect to the pretraining corpus (WebText for GPT-2). 

We report results for $\alpha=2.0$, chosen after observing the tradeoff between detoxification and fluency,
but show results for other values of $\alpha$ in Appendix \ref{sec:additional_results}.

\subsection{Evaluation}
\subsubsection{Generation Prompts}
To evaluate the problem of toxic degeneration where a user might unexpectedly receive harmful output from a model, we use a random sample of 10K nontoxic prompts from the RealToxicityPrompts dataset \cite{gehman-etal-2020-realtoxicityprompts}. 

\subsubsection{Baselines}
\paragraph{Domain-adaptive pretraining \cite[DAPT;][]{gururangan-etal-2020-dont}} We further pretrain the base model on the non-toxic subset of OpenWebText. This dataset is obtained by scoring the full OpenWebText corpus with the toxicity classifier from Perspective API\footnote{\url{https://github.com/conversationai/perspectiveapi}} and keeping the least toxic 2 percent of documents, a corpus of about 150K documents, or 63M tokens, following the implementation of this baseline from \citet{gehman-etal-2020-realtoxicityprompts}.

\paragraph{Plug-and-play language models \cite[PPLM;][]{pplm2020}} PPLM uses gradients from a toxicity classifier to update the LM's hidden representations. 
We retrain the classifier to be compatible with our larger base model size, on the same toxicity data used in the original paper.\footnote{\url{https://bit.ly/3yQiCIo}}
Due to the extreme computational expense of PPLM (runtimes are shown in Appendix \ref{subsec:generation_details}), we evaluate PPLM on a random subset of 1K prompts.

\paragraph{Generative discriminators \cite[GeDi;][]{gedi2020}} GeDi uses a class-conditioned LM to provide classification probabilities for all possible next tokens via Bayes' rule. 
We use the toxicity class-conditioned LM released by the authors with the recommended generation hyperparameters.

\paragraph{\methodName (anti-only)} 
We also explore an anti-expert-only ablation of \methodName, by reusing the base model as the expert. To be clear, we substitute $\mathbf z_t^+=\mathbf z_t$ in \autoref{eq:1}, so that we have
\begin{equation}
    \tilde P(X_t\mid\boldsymbol x_{<t})=\text{softmax}\left((1+\alpha)\mathbf z_t-\alpha\mathbf z_t^-\right)
\end{equation}
\noindent We use the toxic anti-expert based on GPT-2 Large and the same hyperparameter value $\alpha=2.0$.

\paragraph{Non-Toxic Expert} Finally, we consider generating directly from the non-toxic expert based on GPT-2 Large.

\medskip \noindent
For all baselines, we use nucleus sampling \cite{nucleussampling2020} with $p=0.9$ to generate up to $20$ tokens. 
Note that for our method, nucleus sampling is done as described in \S\ref{method}, by using the nucleus from the base LM. 
Other training and generation details (e.g., hyperparameters) are described in \autoref{sec:modeling_details}. 

\begin{figure*}[t!]
    \centering
    \includegraphics[width=\textwidth]{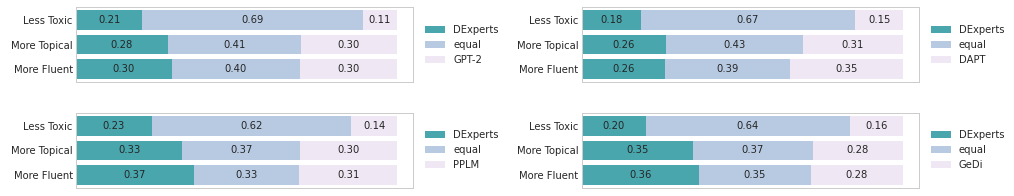}
    \caption{Results of human evaluation for detoxification. \textsc{\methodName} is rated as less toxic more often than every baseline, and equally fluent compared to the base model, GPT-2.}
    \label{fig:toxicity_human_evaluation}
\end{figure*}

\subsubsection{Automatic Evaluation}
We evaluate our generations for toxicity, fluency, and diversity. 
Following previous work \cite{gehman-etal-2020-realtoxicityprompts}, we characterize generation \textbf{toxicity} using the toxicity score from Perspective API, along two axes: 1) the maximum toxicity over $k=25$ generations, and 2) the empirical probability of generating a continuation with toxicity $\geq 0.5$ at least once over $k=25$ generations. 
Generation \textbf{fluency} is measured by the mean perplexity of generated continuations according to a larger pretrained LM, GPT-2 XL. 
Generation \textbf{diversity} is measured using the mean number of distinct $n$-grams, normalized by the length of text \cite{li-etal-2016-diversity}, among the 25 generations for each prompt.
We report Dist-1, Dist-2, and Dist-3 scores for distinct uni-, bi-, and trigrams, respectively. 

\paragraph{Results}
According to automatic metrics shown in Table \ref{tab:toxicity_results}, \methodName substantially outperforms all existing baselines at detoxification.
In particular, \methodName (medium, large) are among the most fluent controllable generation methods, while fully preserving output diversity compared to the base model.
Moreover, the \methodName (anti-only) ablation continues to outperform baselines at detoxification, although with a loss in fluency and diversity that is likely due to the less effective contrast between the base model and anti-expert.
We report the per-generation runtime of each method in Appendix \ref{subsec:generation_details} to demonstrate \methodName's efficiency compared to other decoding-time methods. 
%\nascomment{feels like we're hiding the fact that it's slower than baseline; this language makes it sound like it's peachy. Maybe make that point then point to appendix for more details?}\alisainline{hm... well, it's obviously slower than the baseline, so that would be silly to hide. if I say it's 2-3x the runtime of base model, then I would have to make a more specific claim relative to other methods, too, but there's no easy blanket-statement to make, since \methodName (large) is slower than GeDi. do you think this fix is okay?}

\subsubsection{Human Evaluation}\label{subsubsec:toxicity_human_evaluation}

While automatic toxicity classifiers like Perspective API enable the kind of large-scale evaluation required for systematic comparison of methods, an abundance of work shows that their accuracy is far from ideal \cite{dixon-etal-2018-measuring, sap-etal-2019-risk, davidson-etal-2019-racial, hutchinson-etal-2020-social} in part due to reliance on spurious features, which we discuss in \S\ref{sec:ethics}.
Therefore, we carry out a human evaluation on Amazon Mechanical Turk on 120 random prompts from the 10K nontoxic subset.
For each prompt, we compare four pairs of models: \methodName (large) versus GPT-2 Large, PPLM, DAPT, and GeDi. 
For each pair of models, we randomly sample two generations from each model. 
This results in a total of $120 \text{ prompts}\times 4\frac{\text{pairings}}{\text{prompt}} \times 2\frac{\text{generations}}{\text{pairing}}=960$ comparisons.
Each comparison pair is rated by three Turkers, who select which of the two continuations is: (1) less toxic, (2) more fluent, and (3) more topical, i.e., whether the continuation is natural, relevant, and follows logically from the prompt. 
A screenshot of the user interface is provided in \autoref{sec:human_evaluation}.

\paragraph{Results}
According to human evaluations, \methodName is rated as less toxic more often than all baselines (Figure \ref{fig:toxicity_human_evaluation}). 
In particular, it is rated equally fluent compared to GPT-2, yet less toxic than \mbox{GPT-2} $10\%$ more often than the other way around.
See Appendix \ref{sec:examples} for examples of generations.
% TODO: add more description here. I can't think of anything, though...
% \yejin{this result section reads a bit flat, as it only reports the numbers without offering any insights/conjectures as to \emph{why} Dexperts might perform better than a class conditioned LM or backprop-based PPLM, which at first glance might seem all reasonable methods...}

\subsection{Steering GPT-3}
We next use \methodName to steer GPT-3 Ada. 
Because the OpenAI API\footnote{\url{https://openai.com/api/}} allows access to only the top 100 log probabilities at each time step, we can only modify and sample from the probability distribution over the top 100 tokens.
Nonetheless, results in \autoref{tab:gpt3_results} show that \methodName effectively reduces toxicity from GPT-3 to about the same level as when operating on GPT-2.
This demonstrates that \methodName requires only the output of the base model, and indeed, the (anti-)experts do not need to be built on the base model.
% \nascomment{``be the same model as the base model'' is confusing.  maybe reword as ``be built on the base model''?}\alisainline{is this ok?}

\begin{table}[t!]
    \centering\small
    \begin{tabular}{lcccccc}
    \toprule
        \multirow{2}{*}{\textbf{Model}} &\multicolumn{2}{c}{\textbf{Toxicity} ($\downarrow$)}  \\
        & Avg.~max.~toxicity & Toxicity prob.\\\midrule
        GPT-3 & 0.525 & 0.515 \\
        %\methodName (small) & 0.282 & 0.102 \\
        %\methodName (medium) & 0.288 & 0.093 \\
        \methodName (large) & 0.293 & 0.111 \\
    \bottomrule
    \end{tabular}
    \caption{Results of experiments in detoxifying generations from GPT-3.}%\maarteninline{which dexperts size?}
    \label{tab:gpt3_results}
\end{table}

\subsection{Analysis: Dataset Size}\label{subsec:dataset_size}

\begin{figure}[t!]
    \centering
    \includegraphics[scale=0.4]{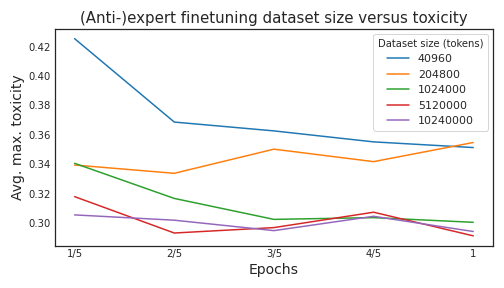}
    \caption{Performance of \methodName when \mbox{(anti-)experts} are trained on differently-sized datasets and evaluated at different checkpoints, calculated on a subset of 1K prompts. For comparison, recall the avg.~max.~toxicity of GPT-2 is 0.527.}
    \label{fig:dataset_size}
\end{figure}

In practice, gathering large amounts of toxic data may be challenging, especially in applications where we would want to customize the anti-expert LM for differing notions of harmful language. 
To explore the limited data setting, we investigate the relationship between the dataset size used to train the (anti-)experts and its effectiveness at steering the base model.
We finetune GPT-2 Large on five different dataset sizes of exactly 40,960, 204.8K, 1.024M, 5.12M, and 10.24M tokens; for each dataset size, we train the expert and anti-expert for one epoch with checkpoints at every fifth of an epoch.
The performance of each ensemble, at every (anti-)expert checkpoint, is show in \autoref{fig:dataset_size}.

We can see that even with a dataset of 40,960 tokens ($\sim$650 comments) corresponding to $<0.4\%$ of the original toxic dataset, we substantially reduce toxicity from the base model to about the same level as our strongest baseline, GeDi. (On one GPU, this corresponds to $\sim$3 minutes of finetuning.) Nonetheless, as the size of the finetuning dataset for (anti-)experts increases, the performance of \methodName increases as well. 

%% file: sections/4.sentiment.tex
\section{Sentiment-Controlled Generation}\label{sec:sentiment}
\begin{table*}[t]
    \centering\small
    \begin{tabular}{llccccccc}
    \toprule
        \multirow{3}{*}{\begin{tabular}{@{}l@{}}\textbf{Target}\\\textbf{Sentiment}\end{tabular}} & \multirow{3}{*}{\textbf{Model}} &\multicolumn{3}{c}{\textbf{\% Positive Sentiment}} & \textbf{Fluency} ($\downarrow$) & \multicolumn{3}{c}{\textbf{Diversity} ($\uparrow$)} \\
        &  & \begin{tabular}{@{}c@{}}Positive\\prompts\end{tabular} & \begin{tabular}{@{}c@{}}Neutral\\prompts\end{tabular} & \begin{tabular}{@{}c@{}}Negative\\prompts\end{tabular}& Output ppl. & Dist-1 & Dist-2 & Dist-3\\\midrule
        \multirow{8}{*}{Positive} & \methodName (large) && 94.46 & 36.42 & 45.83 & 0.56 & 0.83 & 0.83  \\
        & \methodName (medium) & & 94.31 & 33.20 & 43.19 & 0.56 & 0.83 & 0.83  \\
        & \methodName (small) && \textbf{94.57} & 31.64 & 42.08 & 0.56 & 0.83 & 0.84  \\\cmidrule{2-9}
        & GeDi  && 86.01 & 26.80 & 58.41 & 0.57 & 0.80 & 0.79   \\ 
        & Positive expert && 79.83 & \textbf{43.80}& 64.32 & 0.59 & 0.86 & 0.85  \\
        & DAPT  && 77.24 & 14.17 & 30.52 & 0.56 & 0.83 & 0.84  \\
        & \methodName (anti-only) && 60.72 & 4.43 & 46.00 & 0.65 & 0.80 & 0.78  \\ 
        & CTRL  && 61.81 & 18.88 & 43.79 & 0.51 & 0.83 & 0.86  \\
        & PPLM (10\%)  && 52.68 &  8.72 & 142.11 & 0.62 & 0.86 & 0.85 \\\midrule
        
        & GPT-2 & 99.08 & 50.02 & 0.00 & 29.28 & 0.58 & 0.84 & 0.84  \\\midrule
        
        \multirow{8}{*}{Negative} & PPLM (10\%) & 89.74 & 39.05 && 181.78 & 0.63 & 0.87 & 0.86   \\
        & CTRL & 79.05 & 37.63 && 35.94 & 0.50 & 0.83 & 0.86  \\ 
        & \methodName (anti-only) & 93.75 & 34.05 && 44.23 & 0.65 & 0.81 & 0.78  \\
        & DAPT  & 87.43 & 33.28 && 32.86 & 0.58 & 0.85 & 0.84  \\
        & Negative expert & 61.67 & 24.32 && 65.11 & 0.60 & 0.86 & 0.85  \\
        & GeDi  & 39.57 & 8.73 && 84.11 & 0.63 & 0.84 & 0.82  \\\cmidrule{2-9}
        & \methodName (small) & 45.25 & 3.85 && 39.92 & 0.59 & 0.85 & 0.84  \\
        & \methodName (medium) & 40.21 & 3.79 && 43.47 & 0.59 & 0.85 & 0.84  \\
        & \methodName (large) & \textbf{35.99} & \textbf{3.77} && 45.91 & 0.60 & 0.84 & 0.83  \\
    \bottomrule
    \end{tabular}
    \caption{Results for experiments in sentiment-controlled generation. We consider three sets of prompts relative to the base LM: \textbf{neutral prompts}, which are equally likely to lead to positive and negative generations, as well as \textbf{positive prompts} and \textbf{negative prompts}, which lead to overwhelmingly positive and negative generations, respectively. Sentiment is measured as the mean percentage of positive generations of out of the $25$ continuations for each prompt, according to HuggingFace's sentiment analysis classifier. Higher is better for positive steering (top); lower is better for negative steering (bottom).}
    \label{tab:sentiment_results}
\end{table*}

As a second application we consider the well-studied task of controlling the polarity of text's sentiment \cite[e.g.,][]{li-etal-2018-delete, sudhakar-etal-2019-transforming}, steering towards either positive or negative sentiment.
% \nascomment{reworded here; sentiment is not just positive/negative, there's a lot more to it, and we're just looking at polarity}

\subsection{Method}
We use the same pretrained model from \S\ref{sec:toxicity}, GPT-2 Large, as our base LM. We finetune GPT-2 (Small, Medium, Large) on a positive sentiment corpus for our positive LM, and on a negative sentiment corpus for our negative LM. We use Stanford Sentiment Treebank (SST-5; \citealp{socher-etal-2013-recursive}), which contains movie reviews labeled by human raters for sentiment on a scale from 1 (very negative) to 5 (very positive). Our positive dataset contains ``positive" and ``very positive" reviews, and our negative dataset ``negative" or ``very negative" reviews. Each of these sentiment datasets has about 4K reviews.

For ease of notation we consider the positive LM our expert and negative LM our anti-expert, and use $\alpha=\pm 3.2$ for steering in each direction. The tradeoff between fluency and sentiment control for many values of $\alpha$ is shown in \S\ref{subsec:sentiment_analysis}.

\subsection{Evaluation}

\subsubsection{Generation Prompts} 
In order to test our method's ability to control sentiment beyond the domain that the sentiment experts are trained on (movie reviews), we collect a dataset of 100K naturally occurring prompts from the OpenWebText Corpus (OWT) \cite{gokaslan-etal-2019-openwebtext}. Details are outlined in \autoref{sec:sentiment_prompts}. We generate $25$ continuations for each prompt from the base LM, and score them using HuggingFace's  sentiment analysis classifier \cite{wolf-etal-2020-transformers} trained on SST-5 movie reviews. Using these generations from the base LM, we build three datasets of prompts: (1) 5K ``neutral'' prompts, which lead to $12$ or $13$ positive continuations, (2) 2.5K ``negative'' prompts, which lead to $25$ negative continuations, and (3) 2.5K ``positive'' prompts, which lead to $24$ or $25$ positive continuations. We consider the negative and positive prompts \textbf{adversarial settings}, where the task is to steer toward the opposite sentiment of the prompt. 

\subsubsection{Baselines}
We consider the same baselines as in \S\ref{sec:toxicity}, along with a new baseline \cite[CTRL;][]{ctrl2018}.

\paragraph{DAPT} Corresponding to our DAPT baseline in \S\ref{sec:toxicity}, we score all documents in OpenWebText with the HuggingFace sentiment classifier, and keep the most positive 2\% and most negative 2\% (according to the probability of the predicted label) to obtain the positive and negative corpora. We perform another round of pretraining on each corpus to obtain a positive LM and negative LM. 

\paragraph{PPLM} As with toxicity \S\ref{sec:toxicity}, we retrain the sentiment classifier for PPLM with a larger embedding size compatible with our base model. 
The training data used is SST-5. 
Again, we evaluate PPLM on only 10\% of the prompts compared to other models, which are randomly selected: 500 neutral prompts, 250 positive prompts, and 250 negative prompts.

\paragraph{GeDi} We use GeDi with the sentiment class-conditioned LMs released by the original authors, which are trained on IMDB movie reviews \cite{maas-etal-2011-learning}.
(We find that retraining it on SST-5 results in slightly reduced performance, as discussed in Appendix \ref{sec:modeling_details}.)

\paragraph{\methodName (anti-only)} 
To explore whether simply steering away from one sentiment will yield the opposite sentiment, we again explore an anti-expert-only version of \methodName.
As in \S\ref{sec:toxicity}, we reuse the base model as the expert, and use only a negative anti-expert LM for positive steering, and only a positive anti-expert LM for negative steering.
We use $\alpha=\pm 2.0$ for this setting.

\paragraph{Positive/Negative Experts} Again, we consider decoding directly from the corresponding sentiment expert for positive and negative steering.

\paragraph{Conditional Transformer LM \cite[CTRL;][]{ctrl2018}} To control the sentiment of generations from CTRL , we use the ``Reviews'' control code and append a rating of ``5.0'' for positive generations and a rating of ``1.0'' for negative generations. The sentiment training examples for CTRL came from Amazon reviews \cite{amazonreviews2015}.

\medskip \noindent
As with toxicity experiments (\S\ref{sec:toxicity}), we use nucleus sampling with $p=0.9$, and include our training and generation details in Appendix \ref{sec:modeling_details}.

\subsubsection{Automatic Evaluation}
We evaluate our generations for the target sentiment, fluency, and diversity. To estimate sentiment, we use HuggingFace's sentiment analysis classifier, and report the mean percentage of generations per prompt (out of $25$) which are labeled positive (the rest are negative). We evaluate fluency and diversity in the same ways as \S\ref{sec:toxicity}. %: fluency is evaluated as the perplexity of generations according to a larger GPT-2 model, and diversity is evaluated as Dist-$n$ for $n=1,2,3$. 

\paragraph{Results}
As shown in \autoref{tab:sentiment_results}, \methodName greatly outperforms previous controllable generation methods (PPLM, CTRL, DAPT, GeDi) on both neutral prompts and adversarial prompts.
%  \nascomment{this invites me to notice ``positive expert,'' which you address below but presenting this table as an obvious unmitigated win is potentially confusing}\alisainline{oops, this was written before that result. let me think...}
The limited performance of CTRL suggests that the effectiveness of class-conditioned training on domain-specific data is limited to the domain of that data; training on Amazon reviews does not allow generalization outside of the reviews domain.
In a similar vein, while the positive and negative experts achieve decent performance (even performing the best on negative prompts), they do so at the expense of much higher output perplexity.
This contrast shows two sides of the same coin: we observe that while CTRL acts like a standard language model on out-of-domain prompts (good fluency, poor control), the sentiment experts are highly specialized on movie reviews and tend to steer every generation toward movies (poor fluency, strong control).
%\nascomment{did we check qualitatively to see that this was happening?  is there a way to confirm that here? right now it sounds like we're leaping to a conclusion without looking at the data}\alisa{I did look, but I can't think of a good way to demonstrate this}
Meanwhile, DAPT is more effective while maintaining fluency, because its training domain is the same domain as the prompts domain (i.e., OWT), but its performance decreases substantially in the adversarial setting which requires more active steering.
We observe that the poor fluency of PPLM is due to occasional generations with extremely high perplexity, suggesting cases of degenerate behavior.
\methodName with only an anti-expert is mildly effective on neutral prompts (outperforming or matching the performance of CTRL and PPLM), but works very poorly in the adversarial setting, confirming our intuition that steering away from negative sentiment does not provide sufficiently strong guidance for positive sentiment.

% \nascomment{repeating this word in the sentence, reword}
%\nascomment{this hypothesis was not stated, and this term has not been introduced/attached to any of the systems (sorry if I missed it ...)} 
%\nascomment{and the rest?  also, don't use the s-word if you didn't do significance tests; don't do more than one significance test without doing (and explaining that you are doing) multiple hypothesis corrections}

\begin{figure*}[t!]
    \centering
    \includegraphics[width=\textwidth
    ]{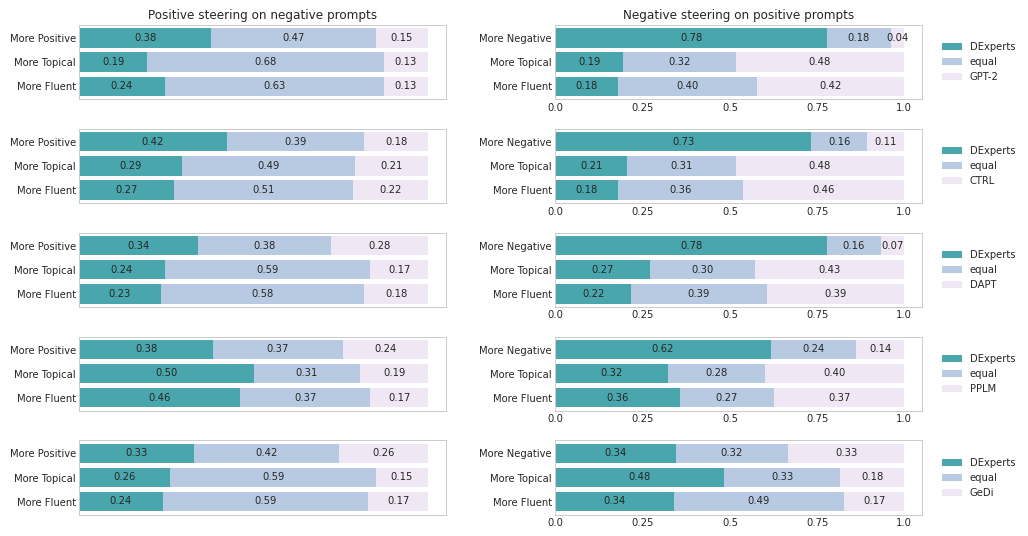}
    \caption{Results of human evaluation for steering toward positivity on negative prompts (left) and steering toward negativity on positive prompts (right). \textsc{\methodName} is substantially more effective at achieving the desired sentiment over every baseline.}
    \label{fig:sentiment_adversarial_human_evaluation}
\end{figure*}

\subsubsection{Human Evaluation}
For human evaluation, we randomly choose 30 neutral prompts, 30 positive prompts, and 30 negative prompts, and consider five pairs of models: \methodName versus GPT-2, CTRL, PPLM, DAPT, and GeDi.
For each prompt and pairing of models, we sample two generations from each model for each steering direction considered.   
This results in a total of $120 \text{ prompts}\times 5\frac{\text{pairings}}{\text{prompt}} \times 2\frac{\text{generations}}{\text{pairing}}=1200$ pairs, each rated by $3$ MTurk workers.
We ask annotators to select which generation achieves the desired sentiment better, along with the fluency and topicality questions from \S\ref{subsubsec:toxicity_human_evaluation}.

\paragraph{Results}
As shown in \autoref{fig:sentiment_adversarial_human_evaluation}, \methodName is substantially more effective at steering toward positivity on negative prompts while achieving better topicality and better fluency compared to all other baselines, including GPT-2.
In the opposite setting of steering toward negativity on positive prompts, the gap in sentiment control performance between \methodName and each of GPT-2, CTRL, DAPT, and PPLM is even more pronounced: \methodName is rated better than its comparison 62--78\% of the time.
While GeDi achieves close to \methodName' performance in this setting, its topicality and fluency are much worse.
The asymmetry, where negative steering appears easier than positive steering for \methodName, is reflected in automatic evaluation as well.
We hypothesize that it is easier to derail a positive prompt with negativity than turn something negative into something positive; but to human readers, these negative continuations may be unexpected \cite[a similar observation was made in previous work;][]{madotto-etal-2020-plug}.
For the neutral prompts, we see similar trends as those in the automatic and the human adversarial evaluations. Due to space constraints, we include those in Appendix \ref{subsec:sentiment_neutral_prompts_human_evaluation}.

\subsection{Analysis: Sentiment versus Fluency}\label{subsec:sentiment_analysis}
\begin{figure}[t!]
    \centering
    \includegraphics[scale=0.38]{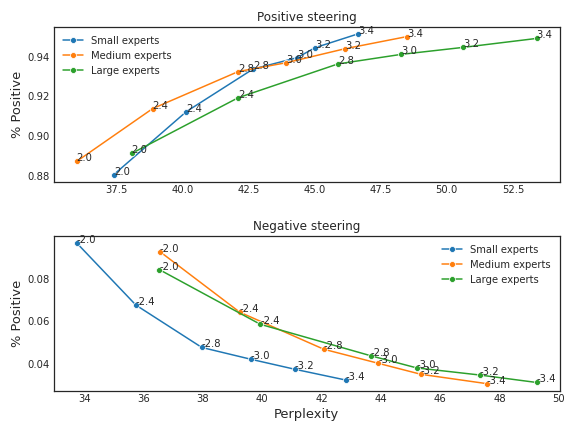}
    \caption{The relationship between output fluency and positivity for different values of $\alpha\in[-3.4,3.4]$. We choose $\alpha=\pm 3.2$ in our experiments. Results are calculated on a subset of 1K neutral prompts.}
    \label{fig:sentiment_hyperparam}
\end{figure}

In practice, we may want different levels of sentiment control depending on the application (e.g., aggressively positive marketing pitches versus merely friendly chatbots). 
\autoref{fig:sentiment_hyperparam} shows the relationship between output sentiment and fluency for different choices of $\alpha\in[-3.4,3.4]$, conditioned on neutral prompts.
The smooth tradeoff suggests that $\alpha$ can by adjusted by a practitioner or user, depending on their application.
In our experiments, we pick $\alpha=\pm 3.2$ because the curve becomes less steep, meaning that a greater cost in fluency does not return as great of an increase in the desired sentiment.
The tradeoff between output toxicity and fluency looks very similar for \methodName detoxification (\S\ref{sec:toxicity}), and is included in Appendix \ref{subsec:toxicity_hyperparam}.

%% file: sections/5.rewriting.tex
\section{Stylistic Rewriting with \methodName}\label{sec:stylistic_rewriting}
As a preliminary exploration, we go beyond generating text continuations to apply \methodName to stylistic rewriting, i.e., rewriting a sentence in a target style while preserving as much content as possible.
We replace the base model with a pretrained autoencoder, BART \cite{lewis-etal-2020-bart}, and use GPT-2 Large sentiment (anti-)experts from \S\ref{sec:sentiment} for steering. 
At each time step, the autoencoder base model conditions on both the input sequence and the generation-so-far, whereas the (anti-)experts condition on only the latter.
As a proof of concept, we show some examples of input/output from this system in \autoref{tab:sentiment_rewriting}.
%\nascomment{strictly speaking, I don't think BART is an autoencoder.  isn't it just a sequence-to-sequence model?}
\begin{table}[h!]
    \centering\small
    \begin{tabular}{@{}p{\columnwidth}@{}}
    \toprule
        Input $\to$ Output Examples \\\midrule
        I love cats and seeing them play with yarn. \\
        $\xrightarrow[]{\alpha=-4.0}$ I love cats and seeing them play with rotten cereal. \\\midrule
        % Corgis are stubborn but very cute. \\
        % $\xrightarrow[]{\alpha=-4.0}$ Corgis are boring idiots. \\\midrule
        Oatmilk is tasty and good for the environment. \\
        $\xrightarrow[]{\alpha=-3.5}$ Oatmilk is toxic and bad for the environment. \\\midrule
        Great food but horrible staff and very very rude workers!\\
        $\xrightarrow[]{\alpha=2.0}$ A very nice restaurant \\
    \bottomrule
    \end{tabular}
    \caption{Examples of input/output from a preliminary system that applies \methodName to stylistic rewriting. Recall $\alpha>0$ indicates positive rewriting, and $\alpha<0$ indicates negative rewriting.}
    \label{tab:sentiment_rewriting}
\end{table}

% We observe that the quality of output is more sensitive to the choice of $\alpha$ and is hindered by the different conditioning given to the sequence-to-sequence base model and language model (anti-)experts. 
This exploration suggests that more innovation is required to apply \methodName to stylistic rewriting, but it is a promising direction. We anticipate future work on the subject.
% \nascomment{reworded here}

%% file: sections/6.related_work.tex
\section{Related Work}\label{sec:related_work}
% \swabha{I second Maarten's high level comments here.}\maarten{commenting this out since I'm rewriting this whole section}\alisa{maarten you're a godsend}
The task of controlling the output of a language generation model has been widely studied by previous work \cite[for a review, see][]{prabhumoye-etal-2020-exploring}.
Prior to using pretrained LMs as a backbone, most work used custom neural models trained for their respective downstream generation tasks, including emotion-aware text generation \cite{ghosh-etal-2017-affect,ficler-goldberg-2017-controlling}, attribute-aware product review generation
\cite{dong-etal-2017-learning}, and friendly or empathetic dialogue response generation
\cite{see-etal-2019-makes, rashkin-etal-2019-towards}.

Since pretrained LMs have shown impressive text generation ability \cite{radford2018improving,gpt22019}, two directions have emerged to control their language generation: training approaches and decoding-time approaches.
Training approaches include finetuning the pretrained LMs on datasets that contain the desired attributes \cite{gururangan-etal-2020-dont} as well as creating a class-conditioned pretrained LM trained on text with specific attributes control code prefixes \cite{ctrl2018}.
In contrast to our method, such approaches can only steer \textit{towards} desired text attributes, they cannot steer \textit{away} from them.
Additionally, training approaches require significant computational resources, which may no longer be feasible with the size of more recent pretrained LMs \cite{gpt32020, Fedus2021SwitchTransformers}.

Decoding-time methods, a more lightweight approach,  have been used controlling the attributes of generated text, as well as for improving its quality \cite{li-etal-2016-diversity,holtzman-etal-2018-learning, welleck-etal-2020-neural}.
PPLM \cite{pplm2020} is a steering method that updates a pretrained model's hidden representations according to the gradient of a classifier with respect to the desired class. 
Unfortunately, this approach is computationally expensive, as shown in this and previous work \cite{gehman-etal-2020-realtoxicityprompts}.
Contemporaneous with our work, FUDGE \cite{yang-klein-2021-fudge} trains classifiers on partial sequences to predict whether an attribute will be satisfied \textit{in the future}, and uses Bayesian factorization to obtain the attribute-conditioned probability distribution.
GeDi \cite{gedi2020} uses Bayes' rule similarly, but computes classification probabilities using the output of class-conditioned LMs rather than directly training a classifier. 
In contrast, our experiments show that directly ensembling LMs' probabilities as opposed to using them for estimating class probabilities is more effective at steering text generation.
%Moreover, \methodName is able to steer away from attributes (e.g., toxicity) without training on examples with the desired attribute (e.g., non-toxic data).

%% file: sections/7.conclusion.tex
\section{Conclusion}\label{sec:conclusion}

We present \textsc{\methodName}, a method for controlled text generation that reweights the predictions of language models based on expert (and anti-expert) opinions.
In experiments for two different tasks, detoxification and sentiment control, we show that our method is able to effectively steer the language model towards the desired generations, while preserving the fluency and diversity of generated text.
As applications built on language models become ubiquitous, \methodName demonstrates promise in steering these models toward safe and user-friendly generations.

%% file: sections/8.ethics_social_impact.tex
\section{Broader Impact and Ethical Implications}\label{sec:ethics}
Our study is motivated by the potential harms of using pretrained language models \cite{bender2021dangers}, specifically their tendency to generate hateful, offensive, or toxic content \cite{sheng-etal-2020-towards,gehman-etal-2020-realtoxicityprompts}.
Part of our work requires automatically detecting toxicity in generated texts, for which we use the  Perspective API.\footnote{\url{https://github.com/conversationai/perspectiveapi}} a commercially deployed toxicity detection tool. 
However, the mismatch between the \textit{construct} of toxicity and its \textit{operationalization} through an automatic classifier can cause biased or unintended model behavior \cite{jacobs-wallach-2021-measurement}.
Specifically, recent work has shown that such hate speech classifiers overestimate the prevalence of toxicity in text that contains a minority identity mention \cite{hutchinson-etal-2020-social, dixon-etal-2018-measuring} or text written by racial minorities \cite{sap-etal-2019-risk, davidson-etal-2019-racial}, therefore having the real possibility of backfiring against its very aim of fairness and inclusive dialogue.
% This mismatch between the \textit{construct} of toxicity and its \textit{operationalization} \cite{jacobs-wallach-2021-measurement} is a limitation of our automatic evaluation. 
To address this limitation, we also perform a \textit{human evaluation} of toxicity, for which we obtained IRB approval and sought to pay our workers a fair wage ($\sim$US\$7--9/h).

We also acknowledge that any controllable detoxification method runs the risk of dual use \cite{Pandya2019DualUse}, specifically, this technology could be used to automatically generate hateful text \cite[e.g., extremist texts;][]{mcguffie2020radicalization}.
For a broader discussion of such risks, and of the risks of large pretrained LMs in general, please see \citet{bender2021dangers}.

Nevertheless, toxicity in pretrained LMs is an unsolved issue \cite{sheng-etal-2019-woman,gehman-etal-2020-realtoxicityprompts}. 
Therefore, we hope future work continues to better define and evaluate the presence of harmful language \cite[e.g.,][]{sap-etal-2020-social}, and to develop systems for mitigating such language that can be personalized to users' diverse experiences with language \cite[e.g., dealing with reclaimed slurs appropriately;][]{Croom2013slurs}.
%, as we explore in our preliminary investigations (in Appendix \ref{subsec:toxicity_customization}).

% \nascomment{nit:  references are inconsistent about initials in names (sometimes there's a period, sometimes not); journals not always capitalized as they should be, some titles with question marks are missing them, some papers have publishers and others don't (I usually only include for books), Gokaslan has no bib info other than title and authors and year; some use very long form names of conferences, others short (I prefer short), NLTK and CTRL and GPT-3 not capitalized.  list of standard things to check for at the end of this doc:  \url{https://docs.google.com/document/d/1T8NiSVUT8nXigsxo9dkMTUhvWRqQ2o6SSZLnFlGMR2I/edit}}

%% file: __main.bbl
\begin{thebibliography}{47}
\expandafter\ifx\csname natexlab\endcsname\relax\def\natexlab#1{#1}\fi

\bibitem[{Bender et~al.(2021)Bender, Gebru, McMillan-Major, and
  Shmitchell}]{bender2021dangers}
Emily Bender, Timnit Gebru, Angelina McMillan-Major, and Shmargaret Shmitchell.
  2021.
\newblock \href {https://dl.acm.org/doi/10.1145/3442188.3445922} {On the
  dangers of stochastic parrots: Can language models be too big?}
\newblock In \emph{Proceedings of the 2021 ACM Conference on Fairness,
  Accountability, and Transparency (FAccT)}.

\bibitem[{Bird and Loper(2004)}]{bird-loper-2004-nltk}
Steven Bird and Edward Loper. 2004.
\newblock \href {https://www.aclweb.org/anthology/P04-3031} {{NLTK}: The
  natural language toolkit}.
\newblock In \emph{Proceedings of the {ACL} Interactive Poster and
  Demonstration Sessions}.

\bibitem[{Brockman et~al.(2020)Brockman, Murati, and Welinder}]{openAI2020API}
Greg Brockman, Mira Murati, and Peter Welinder. 2020.
\newblock \href {https://openai.com/blog/openai-api/} {{OpenAI API}}.
\newblock Blog post.

\bibitem[{Brown et~al.(2020)Brown, Mann, Ryder, Subbiah, Kaplan, Dhariwal,
  Neelakantan, Shyam, Sastry, Askell, Agarwal, Herbert-Voss, Kr{\"u}ger,
  Henighan, Child, Ramesh, Ziegler, Wu, Winter, Hesse, Chen, Sigler, Litwin,
  Gray, Chess, Clark, Berner, McCandlish, Radford, Sutskever, and
  Amodei}]{gpt32020}
T.~Brown, B.~Mann, Nick Ryder, Melanie Subbiah, J.~Kaplan, Prafulla Dhariwal,
  Arvind Neelakantan, Pranav Shyam, Girish Sastry, Amanda Askell, Sandhini
  Agarwal, Ariel Herbert-Voss, G.~Kr{\"u}ger, T.~Henighan, R.~Child, Aditya
  Ramesh, D.~Ziegler, Jeffrey Wu, Clemens Winter, Christopher Hesse, Mark Chen,
  E.~Sigler, Mateusz Litwin, Scott Gray, Benjamin Chess, J.~Clark, Christopher
  Berner, Sam McCandlish, A.~Radford, Ilya Sutskever, and Dario Amodei. 2020.
\newblock \href
  {https://proceedings.neurips.cc/paper/2020/file/1457c0d6bfcb4967418bfb8ac142f64a-Paper.pdf}
  {Language models are few-shot learners}.
\newblock In \emph{Proceedings of the 34th Conference on Neural Information
  Processing Systems (NeurIPS)}.

\bibitem[{Croom(2013)}]{Croom2013slurs}
Adam~M Croom. 2013.
\newblock \href {https://philarchive.org/archive/CROHTD} {How to do things with
  slurs: Studies in the way of derogatory words}.
\newblock In \emph{Language \& {communication}}.

\bibitem[{Dathathri et~al.(2020)Dathathri, Madotto, Lan, Hung, Frank, Molino,
  Yosinski, and Liu}]{pplm2020}
Sumanth Dathathri, Andrea Madotto, Janice Lan, Jane Hung, Eric Frank, Piero
  Molino, Jason Yosinski, and Rosanne Liu. 2020.
\newblock \href {https://openreview.net/pdf?id=H1edEyBKDS} {Plug and play
  language models: A simple approach to controlled text generation}.
\newblock In \emph{Proceedings of the 2020 International Conference on Learning
  Representations (ICLR)}.

\bibitem[{Davidson et~al.(2019)Davidson, Bhattacharya, and
  Weber}]{davidson-etal-2019-racial}
Thomas Davidson, Debasmita Bhattacharya, and Ingmar Weber. 2019.
\newblock \href {https://www.aclweb.org/anthology/W19-3504} {Racial bias in
  hate speech and abusive language detection datasets}.
\newblock In \emph{Proceedings of the Third Workshop on Abusive Language
  Online}.

\bibitem[{Dixon et~al.(2018)Dixon, Li, Sorensen, Thain, and
  Vasserman}]{dixon-etal-2018-measuring}
Lucas Dixon, John Li, Jeffrey Sorensen, Nithum Thain, and Lucy Vasserman. 2018.
\newblock \href
  {https://www.aies-conference.com/2018/contents/papers/main/AIES_2018_paper_9.pdf}
  {Measuring and mitigating unintended bias in text classification}.
\newblock In \emph{Proceedings of the 2018 AAAI/ACM Conference on AI, Ethics,
  and Society (AIES)}.

\bibitem[{Dong et~al.(2017)Dong, Huang, Wei, Lapata, Zhou, and
  Xu}]{dong-etal-2017-learning}
Li~Dong, Shaohan Huang, Furu Wei, Mirella Lapata, Ming Zhou, and Ke~Xu. 2017.
\newblock \href {https://www.aclweb.org/anthology/E17-1059} {Learning to
  generate product reviews from attributes}.
\newblock In \emph{Proceedings of the 15th Conference of the {E}uropean Chapter
  of the Association for Computational Linguistics (EACL)}.

\bibitem[{Fan et~al.(2018)Fan, Lewis, and Dauphin}]{fan-etal-2018-hierarchical}
Angela Fan, Mike Lewis, and Yann Dauphin. 2018.
\newblock \href {https://doi.org/10.18653/v1/P18-1082} {Hierarchical neural
  story generation}.
\newblock In \emph{Proceedings of the 56th Annual Meeting of the Association
  for Computational Linguistics (ACL)}.

\bibitem[{Fedus et~al.(2021)Fedus, Zoph, and
  Shazeer}]{Fedus2021SwitchTransformers}
William Fedus, Barret Zoph, and Noam Shazeer. 2021.
\newblock \href {https://arxiv.org/abs/2101.03961} {{Switch Transformers}:
  Scaling to trillion parameter models with simple and efficient sparsity}.
\newblock {arXiv}.

\bibitem[{Ficler and Goldberg(2017)}]{ficler-goldberg-2017-controlling}
Jessica Ficler and Yoav Goldberg. 2017.
\newblock \href {https://www.aclweb.org/anthology/W17-4912} {Controlling
  linguistic style aspects in neural language generation}.
\newblock In \emph{Proceedings of the Workshop on Stylistic Variation}.

\bibitem[{Fiske(1993)}]{fiske-1993-controlling}
Susan~T Fiske. 1993.
\newblock Controlling other people: the impact of power on stereotyping.
\newblock \emph{{American Psychologist}}.

\bibitem[{Gehman et~al.(2020)Gehman, Gururangan, Sap, Choi, and
  Smith}]{gehman-etal-2020-realtoxicityprompts}
Samuel Gehman, Suchin Gururangan, Maarten Sap, Yejin Choi, and Noah~A. Smith.
  2020.
\newblock \href {https://doi.org/10.18653/v1/2020.findings-emnlp.301}
  {{R}eal{T}oxicity{P}rompts: Evaluating neural toxic degeneration in language
  models}.
\newblock In \emph{Findings of the Association for Computational Linguistics
  (EMNLP Findings)}.

\bibitem[{Ghosh et~al.(2017)Ghosh, Chollet, Laksana, Morency, and
  Scherer}]{ghosh-etal-2017-affect}
Sayan Ghosh, Mathieu Chollet, Eugene Laksana, Louis-Philippe Morency, and
  Stefan Scherer. 2017.
\newblock \href {https://doi.org/10.18653/v1/P17-1059} {Affect-{LM}: A neural
  language model for customizable affective text generation}.
\newblock In \emph{Proceedings of the 55th Annual Meeting of the Association
  for Computational Linguistics (ACL)}.

\bibitem[{Gokaslan and Cohen(2019)}]{gokaslan-etal-2019-openwebtext}
Aaron Gokaslan and Vanya Cohen. 2019.
\newblock \href {https://Skylion007.github.io/OpenWebTextCorpus} {Openwebtext
  corpus}.

\bibitem[{Gururangan et~al.(2020)Gururangan, Marasovi{\'c}, Swayamdipta, Lo,
  Beltagy, Downey, and Smith}]{gururangan-etal-2020-dont}
Suchin Gururangan, Ana Marasovi{\'c}, Swabha Swayamdipta, Kyle Lo, Iz~Beltagy,
  Doug Downey, and Noah~A. Smith. 2020.
\newblock \href {https://doi.org/10.18653/v1/2020.acl-main.740} {Don{'}t stop
  pretraining: Adapt language models to domains and tasks}.
\newblock In \emph{Proceedings of the 58th Annual Meeting of the Association
  for Computational Linguistics (ACL)}.

\bibitem[{Hinton(2002)}]{poe2002}
Geoffrey~E. Hinton. 2002.
\newblock \href
  {https://direct.mit.edu/neco/article/14/8/1771/6687/Training-Products-of-Experts-by-Minimizing}
  {Training products of experts by minimizing contrastive divergence}.
\newblock In \emph{Neural Computation}.

\bibitem[{Holtzman et~al.(2018)Holtzman, Buys, Forbes, Bosselut, Golub, and
  Choi}]{holtzman-etal-2018-learning}
Ari Holtzman, Jan Buys, Maxwell Forbes, Antoine Bosselut, David Golub, and
  Yejin Choi. 2018.
\newblock \href {https://doi.org/10.18653/v1/P18-1152} {Learning to write with
  cooperative discriminators}.
\newblock In \emph{Proceedings of the 56th Annual Meeting of the Association
  for Computational Linguistics (ACL)}.

\bibitem[{Holtzman et~al.(2020)Holtzman, Buys, Forbes, and
  Choi}]{nucleussampling2020}
Ari Holtzman, Jan Buys, Maxwell Forbes, and Yejin Choi. 2020.
\newblock \href {https://openreview.net/forum?id=rygGQyrFvH} {The curious case
  of neural text degeneration}.
\newblock In \emph{Proceedings of the Eighth International Conference on
  Learning Representations (ICLR)}.

\bibitem[{Hutchinson et~al.(2020)Hutchinson, Prabhakaran, Denton, Webster,
  Zhong, and Denuyl}]{hutchinson-etal-2020-social}
Ben Hutchinson, Vinodkumar Prabhakaran, Emily Denton, Kellie Webster, Yu~Zhong,
  and Stephen Denuyl. 2020.
\newblock \href {https://doi.org/10.18653/v1/2020.acl-main.487} {Social biases
  in {NLP} models as barriers for persons with disabilities}.
\newblock In \emph{Proceedings of the 58th Annual Meeting of the Association
  for Computational Linguistics (ACL)}.

\bibitem[{Jacobs and Wallach(2021)}]{jacobs-wallach-2021-measurement}
Abigail~Z. Jacobs and Hannah Wallach. 2021.
\newblock \href {https://dl.acm.org/doi/10.1145/3442188.3445901} {Measurement
  and fairness}.
\newblock In \emph{Proceedings of the 2021 ACM Conference on Fairness,
  Accountability, and Transparency (FAccT)}.

\bibitem[{Keskar et~al.(2019)Keskar, McCann, Varshney, Xiong, and
  Socher}]{ctrl2018}
Nitish~Shirish Keskar, Bryan McCann, Lav Varshney, Caiming Xiong, and Richard
  Socher. 2019.
\newblock \href {https://arxiv.org/abs/1909.05858} {{CTRL}: A conditional
  transformer language model for controllable generation}.
\newblock {arXiv}.

\bibitem[{Krause et~al.(2020)Krause, Gotmare, McCann, Keskar, Joty, Socher, and
  Rajani}]{gedi2020}
Ben Krause, Akhilesh~Deepak Gotmare, Bryan McCann, Nitish~Shirish Keskar,
  Shafiq Joty, Richard Socher, and Nazneen~Fatema Rajani. 2020.
\newblock \href {https://arxiv.org/abs/2009.06367} {{GeDi}: Generative
  discriminator guided sequence generation}.
\newblock {arXiv}.

\bibitem[{Lakoff(1973)}]{lakoff-1973-language}
Robin Lakoff. 1973.
\newblock Language and woman’s place.
\newblock \emph{{Language in Society}}.

\bibitem[{Lewis et~al.(2020)Lewis, Liu, Goyal, Ghazvininejad, Mohamed, Levy,
  Stoyanov, and Zettlemoyer}]{lewis-etal-2020-bart}
Mike Lewis, Yinhan Liu, Naman Goyal, Marjan Ghazvininejad, Abdelrahman Mohamed,
  Omer Levy, Veselin Stoyanov, and Luke Zettlemoyer. 2020.
\newblock \href {https://doi.org/10.18653/v1/2020.acl-main.703} {{BART}:
  Denoising sequence-to-sequence pre-training for natural language generation,
  translation, and comprehension}.
\newblock In \emph{Proceedings of the 58th Annual Meeting of the Association
  for Computational Linguistics (ACL)}.

\bibitem[{Li et~al.(2016)Li, Galley, Brockett, Gao, and
  Dolan}]{li-etal-2016-diversity}
Jiwei Li, Michel Galley, Chris Brockett, Jianfeng Gao, and Bill Dolan. 2016.
\newblock \href {https://doi.org/10.18653/v1/N16-1014} {A diversity-promoting
  objective function for neural conversation models}.
\newblock In \emph{Proceedings of the 2016 Conference of the North {A}merican
  Chapter of the Association for Computational Linguistics (NAACL)}.

\bibitem[{Li et~al.(2018)Li, Jia, He, and Liang}]{li-etal-2018-delete}
Juncen Li, Robin Jia, He~He, and Percy Liang. 2018.
\newblock \href {https://doi.org/10.18653/v1/N18-1169} {Delete, retrieve,
  generate: a simple approach to sentiment and style transfer}.
\newblock In \emph{Proceedings of the 2018 Conference of the North {A}merican
  Chapter of the Association for Computational Linguistics (NAACL)}.

\bibitem[{Maas et~al.(2011)Maas, Daly, Pham, Huang, Ng, and
  Potts}]{maas-etal-2011-learning}
Andrew~L. Maas, Raymond~E. Daly, Peter~T. Pham, Dan Huang, Andrew~Y. Ng, and
  Christopher Potts. 2011.
\newblock \href {https://www.aclweb.org/anthology/P11-1015} {Learning word
  vectors for sentiment analysis}.
\newblock In \emph{Proceedings of the 49th Annual Meeting of the Association
  for Computational Linguistics (ACL)}.

\bibitem[{Madotto et~al.(2020)Madotto, Ishii, Lin, Dathathri, and
  Fung}]{madotto-etal-2020-plug}
Andrea Madotto, Etsuko Ishii, Zhaojiang Lin, Sumanth Dathathri, and Pascale
  Fung. 2020.
\newblock \href {https://doi.org/10.18653/v1/2020.findings-emnlp.219}
  {Plug-and-play conversational models}.
\newblock In \emph{Findings of the Association for Computational Linguistics
  (EMNLP Findings)}.

\bibitem[{McAuley et~al.(2015)McAuley, Targett, Shi, and van~den
  Hengel}]{amazonreviews2015}
Julian McAuley, Christopher Targett, Qinfeng Shi, and Anton van~den Hengel.
  2015.
\newblock \href {https://dl.acm.org/doi/10.1145/2766462.2767755} {Image-based
  recommendations on styles and substitutes}.
\newblock In \emph{Proceedings of the 38th International ACM SIGIR Conference
  on Research and Development in Information Retrieval (SIGIR)}.

\bibitem[{McGuffie and Newhouse(2020)}]{mcguffie2020radicalization}
Kris McGuffie and Alex Newhouse. 2020.
\newblock \href {https://arxiv.org/abs/2009.06807} {The radicalization risks of
  gpt-3 and advanced neural language models}.
\newblock {arXiv}.

\bibitem[{Pandya(2019)}]{Pandya2019DualUse}
Jayshree Pandya. 2019.
\newblock \href
  {https://www.forbes.com/sites/cognitiveworld/2019/01/07/the-dual-use-dilemma-of-artificial-intelligence/?sh=b5eeb76cf029}
  {The {dual-use} dilemma of artificial intelligence}.
\newblock \emph{Forbes Magazine}.

\bibitem[{Prabhumoye et~al.(2020)Prabhumoye, Black, and
  Salakhutdinov}]{prabhumoye-etal-2020-exploring}
Shrimai Prabhumoye, Alan~W Black, and Ruslan Salakhutdinov. 2020.
\newblock \href {https://www.aclweb.org/anthology/2020.coling-main.1}
  {Exploring controllable text generation techniques}.
\newblock In \emph{Proceedings of the 28th International Conference on
  Computational Linguistics (COLING)}.

\bibitem[{Radford et~al.(2018)Radford, Narasimhan, Salimans, and
  Sutskever}]{radford2018improving}
Alec Radford, Karthik Narasimhan, Tim Salimans, and Ilya Sutskever. 2018.
\newblock \href
  {https://s3-us-west-2.amazonaws.com/openai-assets/research-covers/language-unsupervised/language_understanding_paper.pdf}
  {Improving language understanding by generative pre-training}.
\newblock Preprint.

\bibitem[{Radford et~al.(2019)Radford, Wu, Child, Luan, Amodei, and
  Sutskever}]{gpt22019}
Alec Radford, Jeffrey Wu, Rewon Child, David Luan, Dario Amodei, and Ilya
  Sutskever. 2019.
\newblock \href
  {https://d4mucfpksywv.cloudfront.net/better-language-models/language_models_are_unsupervised_multitask_learners.pdf}
  {Language models are unsupervised multitask learners}.
\newblock Preprint.

\bibitem[{Rashkin et~al.(2019)Rashkin, Smith, Li, and
  Boureau}]{rashkin-etal-2019-towards}
Hannah Rashkin, Eric~Michael Smith, Margaret Li, and Y-Lan Boureau. 2019.
\newblock \href {https://doi.org/10.18653/v1/P19-1534} {Towards empathetic
  open-domain conversation models: A new benchmark and dataset}.
\newblock In \emph{Proceedings of the 57th Annual Meeting of the Association
  for Computational Linguistics (ACL)}.

\bibitem[{Sap et~al.(2019)Sap, Card, Gabriel, Choi, and
  Smith}]{sap-etal-2019-risk}
Maarten Sap, Dallas Card, Saadia Gabriel, Yejin Choi, and Noah~A. Smith. 2019.
\newblock \href {https://doi.org/10.18653/v1/P19-1163} {The risk of racial bias
  in hate speech detection}.
\newblock In \emph{Proceedings of the 57th Annual Meeting of the Association
  for Computational Linguistics (ACL)}.

\bibitem[{Sap et~al.(2020)Sap, Gabriel, Qin, Jurafsky, Smith, and
  Choi}]{sap-etal-2020-social}
Maarten Sap, Saadia Gabriel, Lianhui Qin, Dan Jurafsky, Noah~A. Smith, and
  Yejin Choi. 2020.
\newblock \href {https://doi.org/10.18653/v1/2020.acl-main.486} {Social bias
  frames: Reasoning about social and power implications of language}.
\newblock In \emph{Proceedings of the 58th Annual Meeting of the Association
  for Computational Linguistics}.

\bibitem[{See et~al.(2019)See, Roller, Kiela, and Weston}]{see-etal-2019-makes}
Abigail See, Stephen Roller, Douwe Kiela, and Jason Weston. 2019.
\newblock \href {https://doi.org/10.18653/v1/N19-1170} {What makes a good
  conversation? how controllable attributes affect human judgments}.
\newblock In \emph{Proceedings of the 2019 Conference of the North {A}merican
  Chapter of the Association for Computational Linguistics (NAACL)}.

\bibitem[{Sheng et~al.(2020)Sheng, Chang, Natarajan, and
  Peng}]{sheng-etal-2020-towards}
Emily Sheng, Kai-Wei Chang, Prem Natarajan, and Nanyun Peng. 2020.
\newblock \href {https://doi.org/10.18653/v1/2020.findings-emnlp.291} {Towards
  {C}ontrollable {B}iases in {L}anguage {G}eneration}.
\newblock In \emph{Findings of the Association for Computational Linguistics
  (EMNLP Findings)}.

\bibitem[{Sheng et~al.(2019)Sheng, Chang, Natarajan, and
  Peng}]{sheng-etal-2019-woman}
Emily Sheng, Kai-Wei Chang, Premkumar Natarajan, and Nanyun Peng. 2019.
\newblock \href {https://doi.org/10.18653/v1/D19-1339} {The woman worked as a
  babysitter: On biases in language generation}.
\newblock In \emph{Proceedings of the 2019 Conference on Empirical Methods in
  Natural Language Processing and the 9th International Joint Conference on
  Natural Language Processing (EMNLP-IJCNLP)}.

\bibitem[{Socher et~al.(2013)Socher, Perelygin, Wu, Chuang, Manning, Ng, and
  Potts}]{socher-etal-2013-recursive}
Richard Socher, Alex Perelygin, Jean Wu, Jason Chuang, Christopher~D. Manning,
  Andrew Ng, and Christopher Potts. 2013.
\newblock \href {https://www.aclweb.org/anthology/D13-1170} {Recursive deep
  models for semantic compositionality over a sentiment treebank}.
\newblock In \emph{Proceedings of the 2013 Conference on Empirical Methods in
  Natural Language Processing (EMNLP)}.

\bibitem[{Sudhakar et~al.(2019)Sudhakar, Upadhyay, and
  Maheswaran}]{sudhakar-etal-2019-transforming}
Akhilesh Sudhakar, Bhargav Upadhyay, and Arjun Maheswaran. 2019.
\newblock \href {https://doi.org/10.18653/v1/D19-1322} {{``T}ransforming{''}
  delete, retrieve, generate approach for controlled text style transfer}.
\newblock In \emph{Proceedings of the 2019 Conference on Empirical Methods in
  Natural Language Processing and the 9th International Joint Conference on
  Natural Language Processing (EMNLP-IJCNLP)}.

\bibitem[{Welleck et~al.(2020)Welleck, Kulikov, Roller, Dinan, Cho, and
  Weston}]{welleck-etal-2020-neural}
Sean Welleck, Ilia Kulikov, Stephen Roller, Emily Dinan, Kyunghyun Cho, and
  Jason Weston. 2020.
\newblock \href {https://openreview.net/forum?id=SJeYe0NtvH} {Neural text
  generation with unlikelihood training}.
\newblock In \emph{Proceedings of the Eighth International Conference on
  Learning Representations (ICLR)}.

\bibitem[{Wolf et~al.(2020)Wolf, Debut, Sanh, Chaumond, Delangue, Moi, Cistac,
  Rault, Louf, Funtowicz, Davison, Shleifer, von Platen, Ma, Jernite, Plu, Xu,
  Le~Scao, Gugger, Drame, Lhoest, and Rush}]{wolf-etal-2020-transformers}
Thomas Wolf, Lysandre Debut, Victor Sanh, Julien Chaumond, Clement Delangue,
  Anthony Moi, Pierric Cistac, Tim Rault, Remi Louf, Morgan Funtowicz, Joe
  Davison, Sam Shleifer, Patrick von Platen, Clara Ma, Yacine Jernite, Julien
  Plu, Canwen Xu, Teven Le~Scao, Sylvain Gugger, Mariama Drame, Quentin Lhoest,
  and Alexander Rush. 2020.
\newblock \href {https://doi.org/10.18653/v1/2020.emnlp-demos.6} {Transformers:
  State-of-the-art natural language processing}.
\newblock In \emph{Proceedings of the 2020 Conference on Empirical Methods in
  Natural Language Processing (EMNLP): System Demonstrations}.

\bibitem[{Yang and Klein(2021)}]{yang-klein-2021-fudge}
Kevin Yang and Dan Klein. 2021.
\newblock \href {https://arxiv.org/pdf/2104.05218.pdf} {{FUDGE}: Controlled
  text generation with future discriminators}.
\newblock In \emph{Proceedings of the 2021 Conference of the North {A}merican
  Chapter of the Association for Computational Linguistics (NAACL)}.

\end{thebibliography}
